\documentclass{article}

\usepackage[utf8]{inputenc} 
\usepackage[T1]{fontenc}    

\usepackage{microtype}
\usepackage{graphicx}
\usepackage{subfigure}
\usepackage{booktabs} 

\usepackage{hyperref}

\usepackage{url}            
\usepackage{amsfonts}       
\usepackage{nicefrac}       
\usepackage{microtype}      
\usepackage{color}
\usepackage{amsthm}
\usepackage{amsmath}
\usepackage{amssymb}
\usepackage{placeins}
\usepackage{threeparttable}
\usepackage{bm}
\usepackage{soul}
\usepackage[normalem]{ulem}
\usepackage{pifont}
\usepackage{nicefrac}
\usepackage{cleveref}
\usepackage{paralist}
\usepackage{makecell}
\usepackage{caption}
\usepackage{titling}
\usepackage{authblk}
\usepackage{multirow}
\usepackage{wrapfig}
\usepackage{soul}
\usepackage{hyperref}
\usepackage[colorinlistoftodos]{todonotes}

\usepackage[inline]{enumitem}

\theoremstyle{definition}
\newtheorem{definition}{Definition}[section]

\usepackage[accepted]{icml2020}

\icmltitlerunning{MetaFun: Meta-Learning with Iterative Functional Updates}

\usepackage[acronym,smallcaps,nowarn,section,nogroupskip,nonumberlist]{glossaries}

\glsdisablehyper 

\newacronym{CNP}{cnp}{Conditional Neural Process}
\newacronym{NP}{np}{Neural Process}
\newacronym{ANP}{anp}{Attentive Neural Process}
\newacronym{MAML}{maml}{Model Agnostic Meta Learning}
\newacronym{LEO}{leo}{Latent Embedding Optimisation}
\newacronym{RKHS}{rkhs}{Reproducing kernel Hilbert space}
\newacronym{GP}{gp}{Gaussian Process}

\newacronym{KRR}{krr}{Kernel Ridge Regression}

\newacronym{DP}{dp}{Dot-Product Attention}

\newacronym{KG}{kg}{Kernel Gradient}

\newacronym{DFP}{dfp}{Dot-Product Functional Pooling}
\newcommand{\DFP}[1]{\operatorname{\textsc{dfp}}\left(#1\right)}

\newacronym{KFP}{kfp}{Kernel Functional Pooling}
\newcommand{\KFP}[1]{\operatorname{\textsc{kfp}}\left(#1\right)}

\newacronym{MLP}{mlp}{multi-layer perceptron}
\newcommand{\MLP}[1]{\operatorname{\textsc{mlp}}\left(#1\right)}

\newcommand{\cmark}{\ding{51}}%
\newcommand{\xmark}{\ding{55}}%

\begin{document}

\twocolumn[
\icmltitle{MetaFun: Meta-Learning with Iterative Functional Updates}

\icmlsetsymbol{equal}{*}

\begin{icmlauthorlist}
\icmlauthor{Jin Xu}{stats}
\icmlauthor{Jean-Francois Ton}{stats}
\icmlauthor{Hyunjik Kim}{dm}
\icmlauthor{Adam R. Kosiorek}{stats,rob}
\icmlauthor{Yee Whye Teh}{stats}
\end{icmlauthorlist}

\icmlaffiliation{stats}{Department of Statistics, University of Oxford, Oxford, United Kingdom}
\icmlaffiliation{dm}{Google DeepMind, London, United Kingdom}
\icmlaffiliation{rob}{Applied AI Lab, Oxford Robotics Institute, University of Oxford, Oxford United Kingdom}

\icmlcorrespondingauthor{Jin Xu}{jin.xu@stats.ox.ac.uk}

\icmlkeywords{Meta-learning, Few-shot, Kernel, Attention, Functional gradient descent}

\vskip 0.3in
]

\printAffiliationsAndNotice{} 

\begin{abstract}
We develop a functional encoder-decoder approach to supervised meta-learning, where labeled data is encoded into an infinite-dimensional functional representation rather than a finite-dimensional one. Furthermore, rather than directly producing the representation, we learn a neural update rule resembling functional gradient descent which iteratively improves the representation. The final representation is used to condition the decoder to make predictions on unlabeled data. Our approach is the first to demonstrates the success of encoder-decoder style meta-learning methods like conditional neural processes on large-scale few-shot classification benchmarks such as miniImageNet and tieredImageNet, where it achieves state-of-the-art performance.
\end{abstract}

\def\xx{\textbf{x}}
\def\yy{\textbf{y}}
\def\zz{\textbf{z}}
\def\rr{\textbf{r}}
\def\ff{\textbf{f}}
\def\uu{\textbf{u}}
\def\hh{\textbf{h}}
\def\EE{\mathbb{E}}
\def\CC{\mathbf{C}}
\def\TT{\mathbf{T}}
\def\loss{\mathcal{L}}
\def\reg{\mathcal{R}}
\def\KL{\mathsf{KL}}
\def\ptrain{p_\text{train}}

\section{Introduction} \label{sec:introduction}

The goal of meta-learning is to be able to generalise to new tasks from the same task distribution as the training tasks. In supervised meta-learning, a task can be described as making predictions on a set of unlabelled data points (\emph{target}) by effectively learning from a set of data points with labels (\emph{context}).
Various ideas have been proposed to tackle supervised meta-learning from different perspectives \citep{andrychowicz2016learning,ravi2016optimization,finn2017model,koch2015siamese,snell2017prototypical,vinyals2016matching,santoro2016meta,rusu2018meta}. In this work, we are particularly interested in a family of meta-learning models that use an encoder-decoder pipeline, such as Neural Processes \citep{garnelo2018conditional,garnelo2018neural}. The encoder is a permutation-invariant function on the context that summarises the context into a task representation, while the decoder produces a predictive model for the targets, conditioned on the task representation. The objective of meta-learning is then to learn the encoder and the decoder such that the produced predictive model generalises well to the targets of new tasks.

Previous works in this category such as the \gls{CNP} and the \gls{NP} \citep{garnelo2018conditional,garnelo2018neural} use sum-pooling operations to produce finite-dimensional, vectorial, task representations. In this work, we investigate the idea of summarising tasks with infinite-dimensional functional representations.
Although there is a theoretical guarantee that sum-pooling of instance-wise representations can express any set function (\textit{universality}) \citep{zaheer2017deep,bloem2019probabilistic}, in practice \gls{CNP} and \gls{NP} tend to underfit the context \citep{kim2019attentive}. 
This observation is in line with the theoretical finding that for universality, the dimension of the task representation should be at least as large as the cardinality of the context set, if the encoder is a smooth function \citet{wagstaff2019limitations}. 
We develop a method that explicitly uses functional representations. Here the effective dimensionality of the task representation grows with the number of context points, which addresses the aforementioned issues of fixed-dimensional representations.
Moreover, in practice it is difficult to model interactions between data points with only sum-pooling operations. This issue can be partially addressed by inserting modules such as relation networks \citep{sung2018learning,rusu2018meta} or set transformers \citep{lee2019set} before sum-pooling. However, only within-context but not context-target interactions can be modelled by these modules.
The construction of our functional representation involves measuring similarities between all data points, which naturally contains information regarding interactions between elements in either the context or the target.

Furthermore, rather than producing the functional representation in a single pass, we develop an approach that \textit{learns} iterative updates to encode the context into the task representation.
In general, learning via iterative updates is often easier than directly learning the final representation, because of the error-correcting opportunity at each iteration. 
For example, an iterative parameterisation of the encoder in Variational Autoencoders (VAEs) has been demonstrated to be effective in reducing the amortisation gap \citep{marino2018iterative}, 
while in meta-learning, both learning to learn methods \citep{andrychowicz2016learning,ravi2016optimization} and \gls{MAML} \citep{finn2017model} use iterative updating procedures to adapt to new tasks, although these update rules operate in parameter space rather than function space.
Therefore, it is reasonable to conjecture that iterative structures are favourable inductive biases for the task encoding process.

In summary,
the primary contribution of this work is a meta-learning approach that learns to summarise a task using a functional representation constructed via iterative updates. 
We apply our approach to solve  meta-learning problems on both regression and classification tasks, and achieve state-of-the-art performance on heavily benchmarked datasets such as miniImageNet \citep{vinyals2016matching} and tieredImageNet \citep{ren2018meta}, which has never been demonstrated with encoder-decoder meta-learning methods without \gls{MAML}-style gradient updates.
We also conducted an ablation study to understand the effects of the different model components.

\section{MetaFun} \label{sec:metafun}

\begin{figure}
  \centering
    \includegraphics[width=0.99\linewidth]{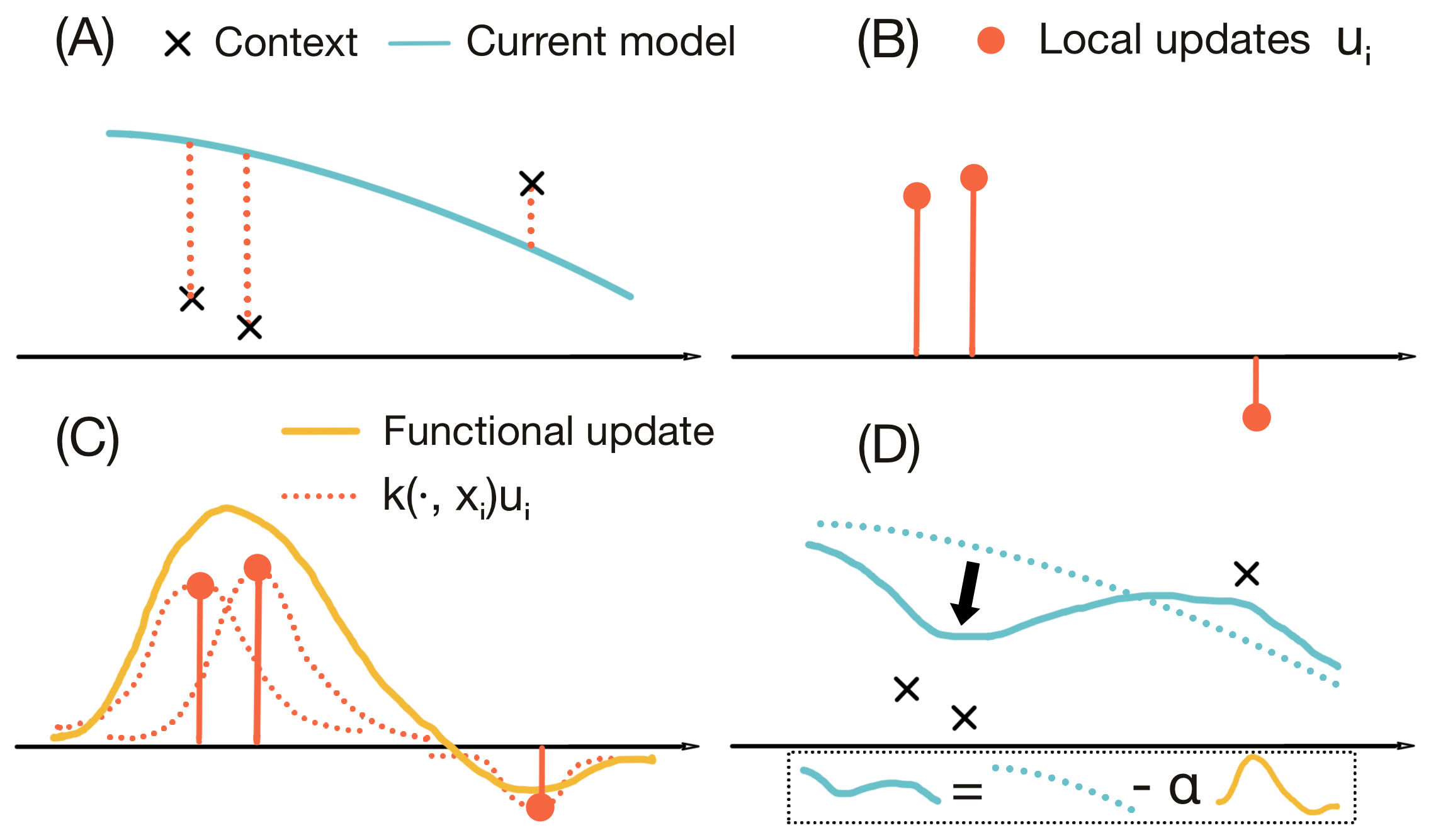}
\caption{To illustrate the iterative procedure in MetaFun, we consider a simpler case where our functional representation is just a predictor for the task. (A) The figure depicts a 1D regression task with the current predictor. (B) Local updates are computed by evaluating the functional representation (the current predictor) on the context inputs, and comparing it to the corresponding context outputs. Here we simply measure differences between evaluations (predictions) and outputs. (C) We apply functional pooling to aggregate local updates into a global functional update, which generalises the local updates to the whole input domain. (D) The functional update is applied to the current functional representation with a learning rate $\alpha$.}
\label{fig:cartoon}
\end{figure}

Meta-learning, or learning to learn, leverages past experiences to quickly adapt to tasks $\mathcal{T}\sim p(\mathcal{T})$ drawn iid from some task distribution.
In supervised meta-learning, a task $\mathcal{T}$ takes the form of $\mathcal{T}=\{ \ell, \{(\xx_i,\yy_i)\}_{i\in \CC}, \{(\xx'_j,\yy'_j)\}_{j\in \TT}\}$, where $\xx_i,\xx'_j\in \mathcal{X}$ are inputs, $\yy_i,\yy'_j\in \mathcal{Y}$ outputs, $\ell$ is the loss function to be minimised, $\{(\xx_i,\yy_i)\}_{i\in \CC}$ is the context, and $\{(\xx'_j,\yy'_j)\}_{j\in \TT}$ is the target. We consider the process of learning as constructing a predictive model using the task context and refer to the mapping from context $\{(\xx_i,\yy_i)\}_{i\in \CC}$ to a predictive model $f = \Phi(\{(\xx_i,\yy_i)\}_{i\in \CC}; \phi)$ as the \emph{learning model} parameterised by $\phi$. In our formulation, the objective of meta-learning is to optimise the learning model such that the expected loss on the target under $f$ is minimised, formally written as:
\begin{align} \label{eq:meta-objective}
    f &= \Phi(\{(\xx_i,\yy_i)\}_{i\in \CC}; \phi) \nonumber \\
    \phi^* &= \underset{\phi}{\arg\min}
    \; \EE_{\mathcal{T} \sim p(\mathcal{T})} \left[ \frac{1}{|\TT|} \sum_{j\in \TT} \ell(f(\xx'_j),\yy'_j) \right]\,,
\end{align}
where both $\{(\xx_i,\yy_i)\}_{i\in \CC}$, $\{(\xx'_j,\yy'_j)\}_{i\in \TT}$ come from task $\mathcal{T}$.

\begin{figure*}
  \centering
    \includegraphics[width=0.99\linewidth]{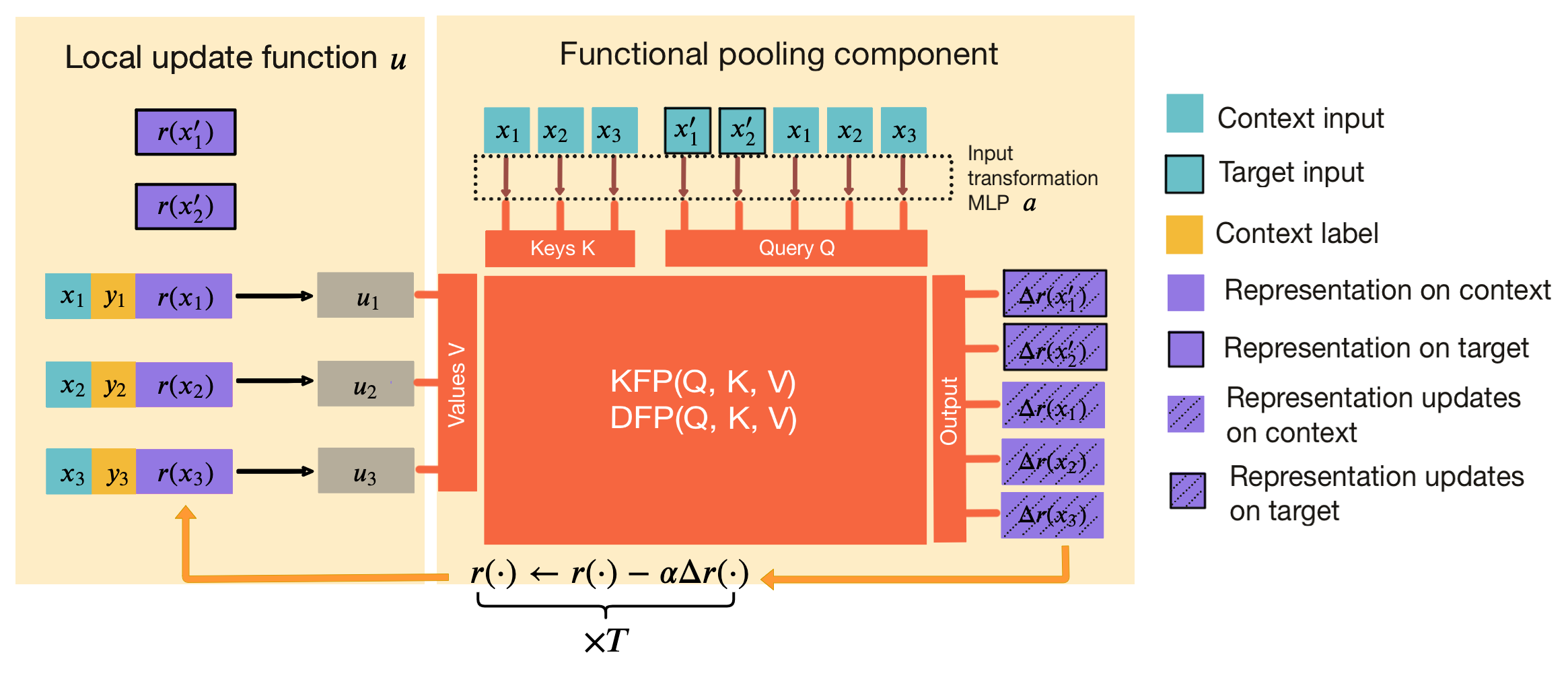}
\caption{This figure illustrates the iterative computation of functional representation in MetaFun. At each iteration, we first evaluate the current functional representation at both context and target points. Then the shared local update function $u$ takes in each context point and the corresponding evaluation as inputs, and produces local update $u_i$. Next, we apply (kernel-based or attention-based) functional pooling  to aggregate local updates $u_i$ into a functional update $\Delta r(\cdot)$, which for each query is a linear combination of local updates $u_i$ weighted by similarities between this query and all keys. Finally, the functional updates are evaluated for both the context and the target, and are applied to the corresponding evaluations of functional representation with a learning rate $\alpha$.}
\label{fig:metafun}
\end{figure*}

\subsection{Learning Functional Task Representation} \label{sub:learning task representation}

Like previous works such as \gls{CNP} and \gls{NP}, we construct the learning model using an encoder-decoder pipeline, where the encoder $\Phi_e(\{(\xx_i,\yy_i)\}_{i\in \CC}; \phi_e)$ is a permutation-invariant function of the context producing a task representation.
In past works, pooling operations are usually used to enforce permutation-invariance. \gls{CNP} and \gls{NP} use sum-pooling: $\rr=\sum_{i\in\CC} \rr_i$, where $\rr_i=h(\xx_i,\yy_i;\phi_e)$ is a representation for context pair $\xx_i,\yy_i$, and $\rr$ is a fixed-dimensional task representation. Instead, we introduce functional-pooling operations, which also enforce permutation-invariance but output a function that can loosely be interpreted as an infinite-dimensional representation.

\theoremstyle{definition}
\begin{definition}[Functional pooling] Let $k(\cdot,\cdot)$ be a real-valued similarity measure, and $\{(\xx_i,\rr_i)\}_{i\in\CC}$ be a set of key-value pairs with $\xx_i\in\mathcal{X}$, $\rr_i\in\mathcal{R}$. \emph{Functional pooling} is a mapping $\operatorname{\textsc{FunPooling}}:(\mathcal{X}\times\mathcal{R})^{|\CC|}\rightarrow \mathcal{H}$ defined as
\begin{align}
\hspace*{-.8em}
    r(\cdot)=\operatorname{\textsc{FunPooling}}(\{(\xx_i,\rr_i)\}_{i\in \CC}) = \sum_{i\in C} k(\cdot,\xx_i) \rr_i\,,
\end{align}
where the output is a function $r:\mathcal{X}\rightarrow \mathcal{R}$ and $\mathcal{H}$ is a space of such functions. 
\end{definition}

In practice, we only need to evaluate this function on a finite query set $\{(\xx'_j,\yy'_j)\}_{j\in\mathcal{Q}}$ (consisting of both contexts and targets; see below). That is, we only need to compute $R=[r(\xx'_1),\ldots,r(\xx'_{|\mathcal{Q}|})]^{\top}$, which can be easily implemented using matrix operations. 
We consider two types of \textsc{FunPooling} here, though others are possible. The kernel-based \textsc{FunPooling} reads as,
\begin{align}
    R=\KFP{Q,K,V} := k_{\text{rbf}}(Q,K) V\,,
\label{eq:kernel_gradient}
\end{align}
where $k_{\text{rbf}}$ is the RBF kernel, $Q=[a(\xx'_1),\ldots,a(\xx'_{|\mathcal{Q}|})]^\top$ is a matrix whose rows are queries,  $a(\cdot)$ is a  transformation mapping inputs into features,  $K=[a(\xx_1),\ldots,a(\xx_{|C|})]^\top$ a matrix whose rows are keys, and $V=[\rr_1,\ldots,\rr_{|C|}]^\top$ a matrix whose rows are values (using terminology from the attention literature). Parameterising input transformation $a$ with a deep neural network can be seen as using deep kernels \citep{wilson2016deep} as the similarity measure.
The second type of \textsc{FunPooling} is given by dot-product attention,
\begin{align}
    R=\DFP{Q,K,V} := \text{softmax}(QK^\top/\sqrt{d_k}) V\,,
\label{eq:attention}
\end{align}
where $d_k$ is the dimension of the query/key vectors.

Our second core idea is that rather than producing the task representation in a single pass (like  previous encoder-decoder meta-learning approaches), we start from an initial representation $r^{(0)}(\cdot)$, and iteratively produce improved representations $r^{(1)}(\cdot),\ldots,r^{(T)}(\cdot)$. 
At each step, a parameterised local update rule $u$ compares $r^{(t)}(\cdot)$ to the context input/output pairs, producing local update values $\uu_i=u(\xx_i,\yy_i,r^{(t)}(\xx_i))$ for each $i\in\CC$. These can then be aggregated into a global update to the task representation using functional pooling,
\begin{align} 
    \uu_i &= u(\xx_i,\yy_i,r^{(t)}(\xx_i))\,, \nonumber \\
    \Delta r^{(t)}(\cdot) &= \textsc{FunPooling}(\{(\xx_i,\uu_i)\}_{i\in\CC})\,, \nonumber \\ 
    r^{(t+1)}(\cdot) &= r^{(t)}(\cdot) - \alpha \Delta r^{(t)}(\cdot)\,,  \label{eq:r-iter}
\end{align}
where $\alpha$ is the step size. 
Once the local update function $u$ and the functional pooling operations are parameterised by neural networks, \Cref{eq:r-iter} defines a  neural update rule operating directly in function space. The functional update $\Delta r^{(t)}(\cdot)$ depends on the current representation $r^{(t)}(\cdot)$ and the context $\{(\xx_i,\yy_i)\}_{i\in \CC}$. \Cref{fig:cartoon} illustrates our iterative procedure in a simplified setting.

The final task representation can then be decoded into a predictor   $f(\cdot)=\Phi_d(r^{(T)}(\cdot); \phi_d)$. 
The specific parametric forms of the decoder take different forms for regression and classification, and are described in \Cref{sub:regression-and-classification}. The decoder requires the evaluation of functional representation $r^{(T)}(\xx)$ at $\xx$ only for predicting $f(\xx)$. Therefore, it is unnecessary to compute the functional representations $r(\cdot)$ (including their functional updates) on all input points. Instead, we compute them only on the context $\{\xx_i\}_{i\in \CC}$ and target inputs $\{\xx'_j\}_{j\in \TT}$. We use $\rr^{(t)}\!=\![r^{(t)}(\xx_1)\ldots r^{(t)}(\xx_{|\CC|}),r^{(t)}(\xx'_1) \ldots r^{(t)}(\xx'_{|\TT|})]^\top$ to denote a matrix where each row is $r^{(t)}(\xx)$ evaluated on either context or target inputs, and let $Q=[a(\xx_1) \ldots a(\xx_{|\CC|}),a(\xx'_1) \ldots a(\xx'_{|\TT|})]^\top$.
\Cref{eq:r-iter} can be implemented using matrix computations as follows,
\begin{align}
\uu_i^{(t)} &= u(\xx_i,\yy_i,\rr^{(t)}_i)\,,\label{eq:value_i_t} \\
U^{(t)} &= [\uu_1^{(t)}, \ldots, \uu_{|\CC|}^{(t)}]^\top\,,\label{eq:values_t}\\
\Delta \rr^{(t)} &= \operatorname{\textsc{kfp}} \text{ or } \DFP{Q, K, U^{(t)}}\,, \label{eq:delta_r}\\
\rr^{(t+1)} &= \rr^{(t)} - \alpha \Delta \rr^{(t)}\, \label{eq:update_r}
\end{align}
where $\rr^{(t)}_i$ denotes the $i$-th row of $\rr^{(t)}$.

To obtain a prediction $\ff_j$ for the target $(\xx'_j,\yy'_j)$, we decode the final representation for this target point: $\ff_j=\Phi_d(\rr^{(T)}_{|\CC|+j}; \phi_d)$, and the overall training loss can be written as:
\begin{equation}
    L(u,a,\phi_d) = \frac{1}{|\TT|} \sum_{j\in \TT} \ell(\ff_j,\yy'_j)\,,
\end{equation}
where the predictions $\ff_j$ depend on $u,a$ and $\phi_d$.

Assuming the width and depth of all our neural network components are bounded by $W$ and $D$ respectively, and the output dimension of $u$ is also less than $W$, the time complexity of our approach is $\mathcal{O}\big(W|\CC|(|\CC|+|\TT|)+W^2 D (|\CC|T+|\TT|)\big)$, and the space complexity is $\mathcal{O}\big((|\CC|+ WT)(|\CC|+|\TT|)+ W^2 D \big)$. For few-shot problems, $|\CC|$ and $|\TT|$ are typically small, and $T \leq 6$ in all our experiments.

\subsection{MetaFun for Regression and Classification} \label{sub:regression-and-classification}

While the proposed framework can be applied to any supervised learning task, the specific parameterisation of learnable components can affect the model performance.
In this section, we specify the parametric forms of our model that work well on regression and classification tasks.
\paragraph{Regression} For regression tasks, we parameterise the local update function $u(\cdot)$ using a multi-layer perceptron\glsunset{MLP} as $u([\xx_i,\yy_i,r(\xx_i)]) = \MLP{[\xx_i,\yy_i,r(\xx_i)]}$, $i \in C$, where $[\cdot]$ is concatenation.
We also use an \gls{MLP} to parametrise the input transformation $a(\cdot)$ in the functional pooling.
The decoder in this case is given by $\mathbf{w}=\MLP{r(\xx)}$, another \gls{MLP}\footnote{
It might be desirable to use other parameterisations of the input transformation $a(\cdot)$, and the decoder $f(\cdot)$, e.g., $f(\xx)~=~\MLP{[\xx,r(\xx)])}$, or feeding $r(\xx)$ to each layer of the \gls{MLP}.
} that outputs $\mathbf{w}$, which then parameterises the predictive model $f=\MLP{\xx;\mathbf{w}}$.

Note that our model can easily be modified to incorporate Gaussian uncertainty by adding an extra output vector for the predictive standard deviation: $P(\yy|\xx)=\mathcal{N}(\mu_{\mathbf{w}}(\xx), \sigma_{\mathbf{w}}(\xx)), \mathbf{w}=\MLP{r(\xx)}$. For further architecture details, see Appendix.

\paragraph{Classification} For $K$-way classification, we divide the latent functional representation $r(\xx)$ into $K$ parts $[r^1(\xx), \dots, r^K(\xx)]$, where $r^k(\xx)$ corresponds to the class $k$. 
Consequently, the local update function $u(\cdot)$ also has $K$ parts, that is, $u([\xx_i,\yy_i,r(\xx_i)]) = [u^1(\cdot), \dots, u^K(\cdot)]$. In this case, $\yy_i=[y_i^1,\dots,y_i^K]$ is the class label expressed as a one-hot vector; the $u^k$ is defined as follows,
\begin{equation} \label{eq:value-function-classification}
    \begin{aligned}
    u^k([\xx_i,\yy_i,r(\xx_i)]) &= y_i^k u_{+}(m(r^k(\xx_i)), \bm{m}_i)\\
    &+ (1-y_i^k) u_{-}(m(r^k(\xx_i)), \bm{m}_i)\,,
    \end{aligned}
\end{equation}
where $\bm{m}_i = \sum_{k=1}^K m(r^k(\xx_i))$ summarises representations of all classes, and $m$, $u_{+}$, $u_{-}$ are parameterised by separate \gls{MLP}s.
With this formulation, we update the class representations using either $u_{+}$ (when the label matches $k$) or $u_{-}$ (when the label is different to $k$), so that labels are not concatenated to the inputs, but directly used to activate different model components, which is crucial for model performance. Furthermore, interactions between data points in classification problems include both within-class and between-class interactions. Our approach is able to integrate two types of interactions by having separate functional representation for each class and computing local updates for each class differently based on class membership of each data point.
In fact, this formulation resembles the structure of the local update rule in functional gradient descent for classification tasks, which is a special case of our approach (see \Cref{sec:related_work}).
Same as in regression tasks, the input transformation $a(\cdot)$ in the functional pooling is still an \gls{MLP}.
The parametric form of the decoder is the same as in \gls{LEO} \citep{rusu2018meta}.
The class representation $r^k(\xx)$ generates weights $\mathbf{w^k}\sim \mathcal{N}(\mu(r^k(\xx)),\sigma(r^k(\xx)))$ where $\mu$ and $\sigma$ are \gls{MLP}s or just linear functions,
and the final prediction is given by
\begin{align}
    P(\yy=k | \xx) = \text{softmax} ( \xx^T \mathbf{w} )_k\,,
\end{align}
where $\mathbf{w} = [\mathbf{w^1}, \dots, \mathbf{w^K}]$, $k=1,\dots,K$.
Hyperparameters of all components are described in Appendix.

\section{Related Work} \label{sec:related_work}

\paragraph{Functional Gradient Descent} Functional gradient descent \citep{mason1999functional,guo2001norm} is an optimisation algorithm used to minimise the objective function by moving in the direction of the negative gradient in function space. 
To ensure smoothness, we may work with functions in a \gls{RKHS} \citep{aronszajn1950theory,berlinet2011reproducing} defined by a kernel $k(\xx,\xx')$.
Given a function $f$ in the \gls{RKHS}, we are interested in minimising the supervised loss $L(f)= \sum_{i\in C} \ell(f(\xx_i),\yy_i)$ with respect to $f$. We can do so by computing the functional derivative and use it to iteratively update $f$ (see Appendix for more details),
\begin{align} 
    \hspace{-0.5em}f^{(t+1)}(\xx) = f^{(t)}(\xx)
    - \alpha \sum_{i\in C} k(\xx,\xx_i) \nabla \ell(f^{(t)}(\xx_i),\yy_i)
    \label{eq:functionupdate}
\end{align}
with step size $\alpha$, and $\nabla \ell(f^{(t)}(\xx_i),\yy_i)$ denotes gradient w.r.t. to predictions in the loss function $\ell$.

The update rule in \Cref{eq:r-iter} becomes that of functional gradient descent in \Cref{eq:functionupdate} when 
\begin{enumerate}[label={\upshape(\roman*)}]
    \item A trivial decoder $f(\xx)=\Phi_d(r(\xx);\phi_d)(\xx)=r(\xx)$ is used, so the functional representation $r(\xx)$ is the same as the predictive model $f(\xx)$.
    \item Kernel functional pooling $\operatorname{\textsc{KFP}}$ is used and the kernel function is fixed.
    \item Using gradient-based local update function $u(\xx,\yy,f(\xx))=\nabla \ell(f(\xx),\yy)$.
\end{enumerate}

Furthermore, for a $K$-way classification problem, we predict $K$-dimensional logits $f(\xx)=[f^{1}(\xx),\dots,f^{K}(\xx)]^\top$, and use cross entropy loss as follows:
\begin{equation} 
    \ell(f(\xx),\yy) = - \sum_{k=1}^K y^k \log \frac{e^{f^{k}(\xx)}}{\sum_{k'=1}^K e^{f^{k'}(\xx)}}\,,
\end{equation}
where $\yy=[y^1,\dots,y^K]^\top$ is the one-hot label for $\xx$.

The gradient-based local update function is now $\nabla \ell(f(\xx),\yy)=[\partial_1 \ell(f(\xx),\yy),\ldots,\partial_K \ell(f(\xx),\yy)]^\top$ where $\partial_k \ell(f(\xx),\yy)$ is partial derivative w.r.t. each predictive logit:
\begin{align} \label{eq:fgd-classification}  
    \partial_k \ell(f(\xx),\yy) &= \frac{e^{f^{k}(\xx)}}{\sum_{k'=1}^K e^{f^{k'}(\xx)}} - y^k\,.
\end{align}
Here $\partial_k \ell(f(\xx),\yy)$ is analogous to $u^k(\cdot)$ in \Cref{eq:value-function-classification}, which is the local update function for class $k$. 

If $m,u_{+},u_{-}$ in \Cref{eq:value-function-classification} are specified rather than being learned, more specifically:
\begin{align}
    m(f^{k'}(\xx)) &= e^{f^{k'}(\xx)}  \nonumber \\ 
    u_{+}(m(f^k(\xx)), \bm{m}) &= \frac{m(f^k(\xx))}{\bm{m}} - 1 \nonumber \\ 
    u_{-}(m(f^k(\xx)), \bm{m}) &= \frac{m(f^k(\xx))}{\bm{m}} \nonumber \\ 
    \bm{m} &= \sum_{k=1}^K m(f^{k'}(\xx))\,,
\end{align}
\Cref{eq:fgd-classification} can be rewritten as:
\begin{align} \label{eq:fgd-classification-rewritten}
\partial_k \ell(f(\xx),\yy) &= y^k u_{+}(m(f^k(\xx)), \bm{m})\nonumber \\
    &+ (1-y^k) u_{-}(m(f^k(\xx)), \bm{m})\,,
\end{align}
which has a similar form as \Cref{eq:value-function-classification}.

Therefore, our approach can be seen as an extension of functional gradient descent, with an additional learning capacity due learnable neural modules which afford more flexibility. From this perspective, our approach tackles supervised meta-learning problems by learning an optimiser in function space.

\paragraph{Supervised Meta-Learning} Various ideas have been proposed to solve the problem of supervised meta-learning. \citet{andrychowicz2016learning,ravi2016optimization} learn the neural optimisers from previous tasks which can be used to optimise models for new tasks. However, these learned optimisers operate in parameter space rather than function space as we do.
\gls{MAML} \citep{finn2017model} learns the initialisation from which models are further adapted for a new task by a few gradient descent steps.
\citet{koch2015siamese,snell2017prototypical,vinyals2016matching} explore the idea of learning a metric space from previous tasks in which data points are compared to each other to make predictions at test time.
\citet{santoro2016meta} demonstrate that Memory-Augmented Neural Networks (MANN) can rapidly integrate the data for a new task into memory, and utilise this stored information to make predictions.

Our approach, in line with previous works such as \gls{CNP} and \gls{NP}, adopt an encoder-decoder pipeline to tackle supervised meta-learning.
The encoder in \gls{CNP} corresponds to a summation of instance-level representations produced by a shared instance encoder.
\Gls{NP}s, on the other hand, use a probabilistic encoder with the same parametric form as \gls{CNP}, but producing a distribution of stochastic representation.
The \gls{ANP} \citep{kim2019attentive} adds a deterministic path in addition to the stochastic path in \gls{NP}. The deterministic path produces a target-specific representation, which can be interpreted as
applying functional pooling (implemented with multihead attention \citep{vaswani2017attention}) to instance-wise representation. However, the representation is directly produced in a single pass rather than iteratively improved as we do, and only regression applications are explored as opposed to few-shot image classification. 
In fact, to achieve high performance for classification tasks, it is crucial for \gls{CNP} to only apply sum-pooling within each class \citep{garnelo2018conditional}, and it is unclear how to follow similar practices in \gls{ANP} with both within-class and between-class interactions still being modelled.
Recently, \citet{gordon2019convolutional} have also extended \gls{CNP} to use functional representations, but for the purpose of incorporating translation equivariance in the inputs as an inductive bias rather than increasing representational capacity as we do. Their approach uses convnets to impose translation equivariance and does not learn a flexible iterative encoder.

Pooling operations are usually used in encoder-decoder meta-learning to enforce permutation invariance in the encoder.
As an example, encoders in both \gls{CNP} and \gls{NP} use simple \emph{sum-pooling} operations. 
More expressive pooling operations have been proposed to model interactions between data points.
\citet{murphy2019janossy} introduces \emph{Janossy pooling} which applies permutation-sensitive functions to all reorderings and averages the outputs, while \citet{lee2019set} use \emph{pooling by multihead attention} (\textsc{pma}), which uses a finite query set to attend to the processed key-value pairs. Loosely speaking, attention-based functional pooling can be seen as having the whole input domain $\mathcal{X}$ as the query set in \textsc{pma}.

\paragraph{Gradient-Based Meta-Learning}
Interestingly, many gradient-based meta-learning methods such as \gls{MAML} can also be cast into an encoder-decoder formulation, because a gradient descent step is a valid permutation-invariant function. For a model $f(\cdot, \theta)$ parameterised by $\theta$, one gradient descent step on the context loss has the following form,
\begin{equation}
    \theta_{t+1} = \theta_t - \alpha \sum_{i\in C} \nabla_{\theta} \ell(f(\xx_i;\theta_t),\yy_i)\,,
\end{equation}
where $\ell$ is the loss function, $\alpha$ is the learning rate, and $\theta_t$ are the model parameters after $t$ gradient steps.
This corresponds to a special case of permutation-invariant functions where we take the instance-wise encoder to be $h_t(\xx_i,\yy_i;\theta_t) = \theta_t/|\CC| - \alpha \nabla_{\theta} \ell(f(\xx_i;\theta_t),\yy_i)$ and apply sum-pooling $\theta_{t+1}=\sum_i h_t(\xx_i,\yy_i;\theta_t)$. 
Multiple gradient-descent steps also result in a permutation-invariant function, which can be proved by induction. We refer to this as a gradient-based encoder.
What follows is that popular meta-learning methods such as \gls{MAML} can be seen as part of the encoder-decoder formulation.
More specifically, in \gls{MAML}, we learn an initialisation of the model parameters $\theta_0$ from training tasks, and adapt to new tasks by running $T$ gradient steps from the learned initialisation. Therefore, $\theta_T$ can be seen as the task representation (albeit very high-dimensional) produced by a gradient-based encoder.
The success of \gls{MAML} on a variety of tasks can be partially explained by the high-dimensional representation and the iterative adaptation by gradient descent, supporting our usage of a functional ('infinite-dimensional') representation and iterative updating procedure. Note, however, that the update rule in \gls{MAML} operates in parameter space rather than function space as in our case. 

Under the same encoder-decoder formulation, a comparison regarding \gls{MAML} and MetaFun can be made, which partially explains why MetaFun can be desirable: Firstly, the updates in \gls{MAML} must lie in its parametric space, while there is no parametric constraint in MetaFun, which is better illustrated in \Cref{fig:sinusoid}. Secondly, \gls{MAML} uses gradient-based updates, while MetaFun uses learned local updates, which potentially contains more information than gradient. Finally, \gls{MAML} does not explicitly consider interactions between data points, while both within-context and context-target interactions are modelled in MetaFun.

\section{Experiments} \label{sec:experiments}

We evaluate our proposed model on both few-shot regression and classification tasks. 
In all experiments that follow, we partition the data into training, validation and test meta-sets, each containing data from disjoint tasks. 
For quantitative results, we train each model with $5$ different random seeds and report the mean and the standard deviation of the test accuracy. For further details on hyperparameter tuning, see the Appendix. All experiments are performed using TensorFlow \citep{tensorflow2015-whitepaper}, and the code is available online \footnote{A tensorflow implementation of our model is available at \url{github.com/jinxu06/metafun-tensorflow}}.

\subsection{1-D Function Regression}

\begin{figure*} 
\begin{minipage}{0.49\textwidth}
    \includegraphics[width=1.0\linewidth]{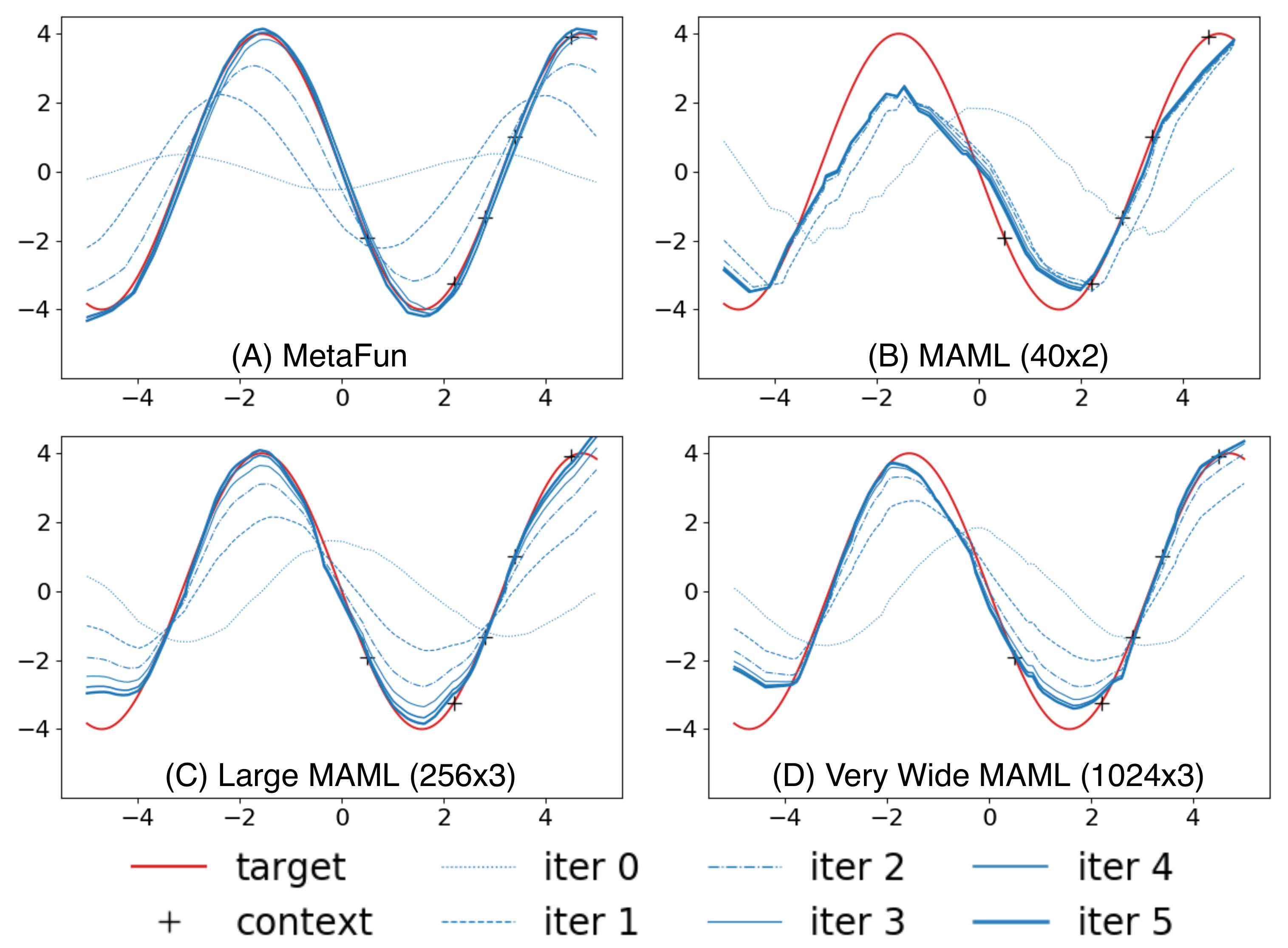}
    \caption{MetaFun is able to learn smooth updates, and recover the ground truth function almost perfectly. While the updates given by \gls{MAML}s are relatively not smooth, especially for \gls{MAML} with less parameters.}
    \label{fig:sinusoid}
\end{minipage} 
\hspace{0.5cm}
\begin{minipage}{0.49\textwidth}
    \includegraphics[width=1.0\linewidth]{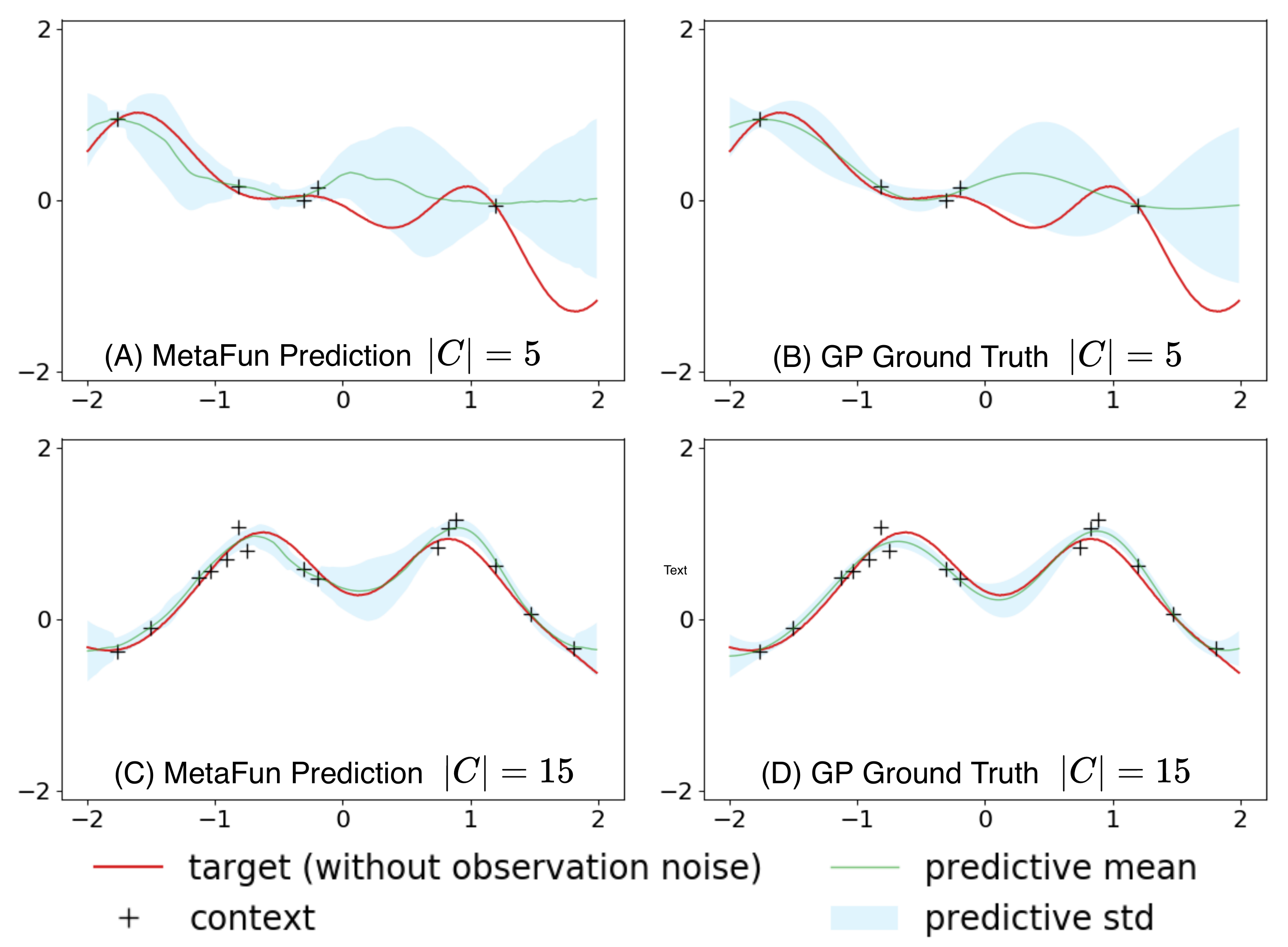}
\caption{Predictive uncertainties for MetaFun matches those for the oracle \gls{GP} very closely in both 5-shot and 15-shot cases. The model is trained on random context size ranging from $1$ to $20$.}
\label{fig:gp}
\end{minipage}
\end{figure*}

We first explore a 1D sinusoid regression task where we visualise the updating procedure in function space, providing intuition for the learned functional updates. Then we incorporate Gaussian uncertainty into the model, and compare our predictive uncertainty against that of a \gls{GP} which generates the data.

\begin{table}[!ht]  
  \centering
  \footnotesize
  \caption{Few-shot regression on sinusoid. \gls{MAML} can beneift from more parameters, but MetaFun still outperforms all \gls{MAML}s despite less parameters being used compared to large \gls{MAML}. We report mean and standard deviation of $5$ independent runs.}
  \begin{tabular}{l|cc}
    \toprule
    Model &  $5$-shot MSE & $10$-shot MSE \\ 
    \midrule
    Original \gls{MAML} & $0.390\pm0.156$ & $0.114\pm0.010$ \\
    Large \gls{MAML} & $0.208\pm0.009$ & $0.061\pm0.004$ \\
    Very Wide \gls{MAML}  & $0.205\pm0.013$ & $0.059\pm0.010$ \\
    MetaFun & $\bm{0.040\pm0.008}$ & $\bm{0.017\pm0.005}$ \\
    \bottomrule
  \end{tabular}
  \label{tb:sinusoid}
\end{table}

\paragraph{Visualisation of functional updates} We train a $T$-step MetaFun with dot-product functional pooling, on a simple sinusoid regression task from \citet{finn2017model}, where each task uses data points of a sine wave. The amplitude $A$ and phase $b$ of the sinusoid varies across tasks and are randomly sampled during training and test time, with $A \in \mathcal{U}(0.1, 5.0)$ and $b \in \mathcal{U}(0, \pi)$. The x-coordinates are uniformly sampled from $\mathcal{U}(-5.0, 5.0)$.
\Cref{fig:sinusoid} shows that our proposed algorithm learns a smooth transition from the initial state to the final prediction at $t=T=5$. Note that although only $5$ context points on a single phase of the sinusoid are given at test time, the final iteration makes predictions close to the ground truth across the whole period.
As a comparison, we use \gls{MAML} as an example of updating in parameter space. The original \gls{MAML} ($40$ units $\times$ $2$ hidden layers) can fit the sinusoid quite well after several iterations from the learned initialisation. However the prediction is not as good, particularly on the left side where there are no context points (see \Cref{fig:sinusoid} B). As we increase the model size to large \gls{MAML} ($256$ units $\times$ $3$ hidden layers), updates become much smoother (\Cref{fig:sinusoid} C) and the predictions are closer to the ground truth. 
We further conduct experiments with a very wide \gls{MAML} ($1024$ units $\times$ $3$ hidden layers), but the performance cannot be further improved (\Cref{fig:sinusoid} D). In \Cref{tb:sinusoid}, we compare the mean squared error averaged across tasks. MetaFun performs much better than all \gls{MAML}s, even though less parameters ($116611$ parameters) are used compared to large \gls{MAML} ($132353$ parameters).

\paragraph{Predictive uncertainties} As another simple regression example, we demonstrate that MetaFun, like \gls{CNP}, can produce good predictive uncertainties. We use synthetic data generated using a \gls{GP} with an RBF kernel and Gaussian observation noise ($\mu=0,\sigma=0.1$), and our decoder produces both predictive means and variances. 
As in \citet{kim2019attentive}, we found that MetaFun-DFP can produce somewhat piece-wise constant mean predictions which is less appealing in this situation. On the other hand, MetaFun-KFP (with deep kernels) performed much better, as can be seen in \Cref{fig:gp}. We consider the cases of $5$ or $15$ context points, and compare our predictions to those for the oracle \gls{GP}. In both cases, our model gave very good predictions. 

\begin{table*}[!htb]  
  \centering
  \footnotesize
  \begin{threeparttable}[]
  \caption{Few-shot Classification Test Accuracy}
  \begin{tabular}{l|cc}
    \toprule
         & \textbf{miniImageNet 5-way}     & \textbf{miniImageNet 5-way}  \\
    Models    & \textbf{1-shot}     & \textbf{5-shot} \\
    \midrule
    \emph{(Without deep residual networks feature extraction):} & & \\
    Matching networks \citep{vinyals2016matching} & $43.56\pm0.84\%$ & $55.31\pm0.73\%$ \\
    Meta-learner LSTM \citep{ravi2016optimization}  & $43.44\pm0.77\%$ & $60.60\pm0.71\%$ \\
    MAML \citep{finn2017model} & $48.70\pm1.84\%$ & $63.11\pm0.92\%$ \\
    LLAMA \citep{grant2018recasting}  & $49.40\pm1.83\%$ & - \\
    REPTILE \citep{nichol2018first}  & $49.97\pm0.32\%$ & $65.99\pm0.58\%$ \\
    PLATIPUS \citep{finn2018probabilistic}  & $50.13\pm1.86\%$ & - \\
    \midrule
    \emph{(Without data augmentation):} & & \\
    Meta-SGD \citep{li2017meta} & $54.24\pm0.03\%$ & $70.86\pm0.04\%$  \\
    SNAIL \citep{mishra2017simple} & $55.71\pm0.99\%$ & $68.88\pm0.92\%$ \\
    \citet{bauer2017discriminative} & $56.30\pm0.40\%$ & $73.90\pm0.30\%$ \\
    \citet{munkhdalai2018rapid} & $57.10\pm0.70\%$ & $70.04\pm0.63\%$ \\
    TADAM \citep{oreshkin2018tadam} & $58.50\pm0.30\%$ & $76.70\pm0.30\%$  \\
    \citet{qiao2018few} & $59.60\pm0.41\%$ & $73.74\pm0.19\%$  \\
    LEO & $\bm{61.76\pm0.08\%}$ & $ 77.59\pm0.12\%$  \\
    MetaFun-DFP & $\bm{62.12\pm0.30\%}$ & $77.78\pm0.12\%$  \\
    MetaFun-KFP & $61.16\pm0.15\%$ & $\bm{78.20\pm0.16\%}$  \\

    \midrule
    \emph{(With data augmentation):} & & \\
    \citet{qiao2018few} & $\bm{63.62\pm0.58\%}$ & $78.83\pm0.36\%$  \\
    LEO & $63.97\pm0.20\%$ & $79.49\pm0.70\%$  \\
    MetaOptNet-SVM \citep{lee2019meta}\tnote{1} & $\bm{64.09\pm0.62\%}$ & $80.00\pm0.45\%$  \\
    MetaFun-DFP & $\bm{64.13\pm0.13\%}$  & 
    $\bm{80.82\pm0.17\%}$  \\
    MetaFun-KFP & $63.39\pm0.15\%$  & 
    $\bm{80.81\pm0.10\%}$  \\
    \bottomrule
  \end{tabular}
  
  \begin{tabular}{l|cc}
    \toprule
     & \textbf{tieredImageNet 5-way}     & \textbf{tieredImageNet 5-way}  \\
    Models    & \textbf{1-shot}     & \textbf{5-shot} \\
    \midrule
    \emph{(Without deep residual networks feature extraction):} & & \\
    MAML \citep{finn2017model} & $51.67\pm1.81\%$ & $70.30\pm0.08\%$ \\
    Prototypical Nets \citep{snell2017prototypical} & $53.31\pm0.89\%$ & $72.69\pm0.74\%$ \\
    Relation Net [in \citet{liu2019learning}] & $54.48\pm0.93\%$ & $71.32\pm0.78\%$ \\
    Transductive Prop. Nets \citep{liu2019learning}\hspace{1cm} & $57.41\pm0.94\%$ & $71.55\pm0.74\%$ \\
    \midrule
    \emph{(With deep residual networks feature extraction):} & & \\
    Meta-SGD & $62.95\pm0.03\%$ & $79.34\pm0.06\%$  \\
    LEO & $66.33\pm0.05\%$ & $81.44\pm0.09\%$  \\
    MetaOptNet-SVM & $65.81\pm0.74\%$ & $81.75\pm0.58\%$  \\
    MetaFun-DFP & $\bm{67.72\pm0.14\%}$ & $82.81\pm0.15\%$  \\
    MetaFun-KFP & $67.27\pm0.20\%$ & $\bm{83.28\pm0.12\%}$  \\
    \bottomrule
  \end{tabular}
  \label{tb:classification}
  \end{threeparttable}
\end{table*}

\subsection{Classification: miniImageNet and tieredImageNet}\label{sub:imagenet}

The \emph{miniImageNet} dataset \citep{vinyals2016matching} consists of 100 classes selected randomly from
the ILSVRC-12 dataset \citep{russakovsky2015imagenet}, and each class contains 600 randomly sampled images. We follow the split in \citet{ravi2016optimization}, where the dataset is divided into training ($64$ classes), validation ($16$ classes), and test ($20$ classes) meta-sets. 
The \emph{tieredImageNet} dataset \citep{ren2018meta} contains a larger subset of the ILSVRC-12 dataset. These classes are further grouped into $34$ higher-level nodes. These nodes are then divided into training ($20$ nodes), validation ($6$ nodes), and test ($8$ nodes) meta-sets. This dataset is considered more challenging because the split is near the root of the ImageNet hierarchy \citep{ren2018meta}. For both datasets, we use the pre-trained features provided by \citet{rusu2018meta}. 

Following the commonly used experimental setting, each few-shot classification task consists of $5$ randomly sampled classes from a meta-set. Within each class, we have either $1$ example ($1$-shot) or $5$ examples ($5$-shot) as context, and $15$ examples as target. For all experiments, hyperparameters are chosen by training on the training meta-set, and comparing target accuracy on the validation meta-set. We conduct randomised hyperparameters search \citep{bergstra2012random}, and the search space is given in Appendix. Then with the model configured by the chosen hyperparameters, we train on the union of the training and validation meta-sets, and report final target accuracy on the test meta-set. 

In \Cref{tb:classification} we compare our approach to other meta-learning methods. The numbers presented are the mean and standard deviation of $5$ independent runs. The table demonstrates that our model outperforms previous state-of-the-art on 1-shot and 5-shot classification tasks for the more challenging tieredImageNet. 
As for miniImageNet, we note that previous work, such as MetaOptNet-SVM \citep{lee2019meta}, used significant data augmentation to regularise their model and hence achieved superior results. For a fair comparison, we also equipped each model with data augmentation and reported accuracy with/without data augmentation. 
However, MetaOptNet-SVM \citep{lee2019meta} uses a different data augmentation scheme involving horizontal flip, random crop, and color (brightness, contrast, and saturation) jitter. On the other hand, MetaFun, \citet{qiao2018few} and LEO \citep{rusu2018meta}, only use image features averaging representation of different crops and their horizontal mirrored versions.
In 1-shot cases, MetaFun matches previous state-of-the-art performance, while in 5-shot cases, we get significantly better results. In \Cref{tb:classification}, results for both MetaFun-DFP (using dot-product attention) and MetaFun-KFP (using deep kernels) are reported. Although both of them demonstrate state-of-the-art performance, MetaFun-KFP generally outperforms MetaFun-DFP for 5-shot problems, but performs slightly worse for 1-shot problems.

\subsection{Ablation Study}

\begin{table*}[!htb] 
  \centering
  \footnotesize
  \begin{threeparttable}[]
  \caption{Ablation Study. We conduct independent randomised hyperparameter search for each number presented, and reported means and standard deviations over 5 independent runs for each.}
  \label{table:few_shot}
  \begin{tabular}{ccc|cc|cc}
    \toprule
    Functional &  Local update &  \multirow{2}{*}{Decoder} &  \multicolumn{2}{c|}{\textbf{MiniImageNet}} & \multicolumn{2}{c}{\textbf{tieredImageNet}} \\ 
     pooling &  function &  & 1-shot   & 5-shot \\
    \midrule
    Attention & NN & \cmark & \bm{$62.12\pm0.30\%$} & $77.78\pm0.12\%$ & \bm{$67.72\pm0.14\%$} & $82.81\pm0.15\%$  \\
    Deep Kernel & NN& \cmark & $61.16\pm0.15\%$ & \bm{$78.20\pm0.16\%$} & $67.27\pm0.20\%$ & \bm{$83.28\pm0.12\%$} \\
    Attention & Gradient & \cmark & $59.63\pm0.19\%$ & $75.84\pm0.04\%$  & $62.55\pm0.10\%$ & $78.18\pm0.09\%$ \\
    Deep Kernel & Gradient & \cmark & $59.73\pm0.21\%$ & $76.41\pm0.14\%$ & $65.24\pm0.11\%$ & $80.31\pm0.16\%$ \\
    \midrule
    SE Kernel & NN & \cmark & $60.04\pm0.19\%$ & $75.25\pm0.12\%$  & $60.81\pm0.30\%$ & $79.70\pm0.20\%$ \\
    Deep Kernel & Gradient & \xmark & $57.67\pm0.16\%$ & $73.55\pm0.04\%$ & $62.53\pm0.17\%$ & $76.86\pm0.07\%$ \\
    \bottomrule
  \end{tabular}
  \label{tb:ablation}
  \end{threeparttable}
\end{table*}

\begin{figure*}
    \includegraphics[width=0.99\linewidth]{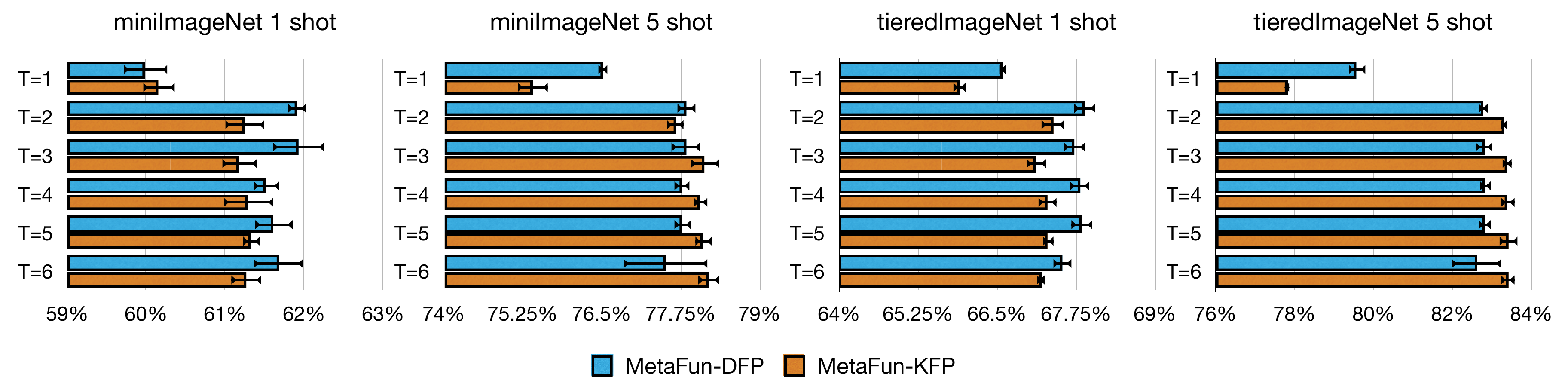}
    \caption{This figure illustrates the accuracy of our approach for varying number of iterations $T=1, \dots,6$, over different few-shot learning problems. For each problem, we use the same configuration of hyperparameters except for the number of iterations and the choice between attention and deep kernels. Error bars (standard deviations) are given by training the same model $5$ times with different random seeds.}
\label{fig:iteration}
\end{figure*}

As stated in \Cref{sub:regression-and-classification}, our model has three learnable components: the local update function, the functional pooling, and the decoder. In this section we explore the effects of using different versions of these components. We also investigate how the model performance would change with different numbers of iterations.

\Cref{tb:ablation} demonstrates that neural network parameterised local update functions, described in \Cref{sub:learning task representation}, consistently outperforms gradient-based local update function, despite the latter having build-in inductive biases. 
Interestingly, the choice between dot-product attention and deep kernel in functional pooling is problem dependent. We found that MetaFun with deep kernels usually perform better than MetaFun with dot product attention on $5$-shot classification tasks, but worse on $1$-shot tasks. We conjecture that the deep kernel is better able to fuse the information across the 5 images per class compared to attention. In the comparative experiments in \Cref{sub:imagenet} we reported results on both.

In addition, we investigate how a simple Squared Exponential (SE) kernel would perform on these few-shot classification tasks. This corresponds to using an identity input transformation function $a$ in deep kernels. \Cref{tb:ablation} shows that using SE kernel is consistently worse than using deep kernels, showing that the heavily parameterised deep kernel is necessary for these problems.

Next, we looked into directly applying functional gradient descent with parameterised deep kernel to these tasks. This corresponds to removing the decoder and using deep kernels and gradient-based local update function (see \Cref{sec:related_work}). Unsurprisingly, this did not fare as well, given as it only has one trainable component (the deep kernel) and the updates are directly applied to the predictions rather than a latent functional representation. 

Finally, \Cref{fig:iteration} illustrates the effects of using different numbers of iterations $T$. On all few-shot classification tasks, we can see that using multiple iterations (two is often good enough) always significantly outperform one iteration. We also note that this performance gain diminishes as we add more iterations. In \Cref{sub:imagenet} we treated the number of iterations as one of the hyperparameters.

\section{Conclusions and Future Work}

In this paper, we propose a novel functional approach for meta-learning called MetaFun. The proposed approach learns to generate a functional task representation and an associated functional update rule, which allows to iteratively update the task representation directly in the function space.
We evaluate MetaFun on both few-shot regression and classification tasks, and
demonstrate that it matches or exceeds previous state-of-the-art results on miniImageNet and tieredImageNet few-shot classification tasks. 

Interesting future research directions include
a) exploring a stochastic encoder and hence working with stochastic functional representations, akin to the \glsreset{NP}\gls{NP},
and b) using local update functions and the functional pooling components whose parameters change with iterations instead of sharing them across iterations, where the added flexibility could lead to further performance gains.

\section*{Acknowledgements}

We would like to thank Jonathan Schwarz for valuable discussion, and the anonymous reviewers for their feedback. Jin Xu and Yee Whye Teh acknowledge funding from Tencent AI Lab through the Oxford-Tencent Collaboration on Large Scale Machine Learning project. Jean-Francois Ton is supported by the EPSRC and MRC through the OxWaSP CDT programme (EP/L016710/1). 

\clearpage
\newpage

\bibliography{metafun}
\bibliographystyle{icml2020}

\onecolumn
\icmltitle{Appendix for \\ MetaFun: Meta-Learning with Iterative Functional Updates}

\appendix 

\section{Functional Gradient Descent}

Functional gradient descent \citep{mason1999functional,guo2001norm} is an iterative optimisation algorithm for finding the minimum of a function. However, the function to be minimised is now a function on functions (\emph{functional}). Formally, a functional $L:\mathcal{H} \rightarrow \mathbf{R}$ is a mapping from a function space $\mathcal{H}$ to a $1D$ Euclidean space $\mathbf{R}$.
Just like gradient descent in parameter space which takes steps proportional to the negative of the gradient, functional gradient descent updates $f$ following the gradient in function space. In this work, we only consider a special function space called \gls{RKHS} (\Cref{sub:rkhs}), and calculate functional gradients in \gls{RKHS} (\Cref{sub:fg}). The algorithm is further detailed in \Cref{sub:fgd-appendix}.

\subsection{Reproducing Kernel Hilbert Space} \label{sub:rkhs}

A Hilbert space $\mathcal{H}$ extends the notion of Euclidean space by introducing inner product $\langle\cdot,\cdot\rangle_{\mathcal{H}}$ which describes the concept of distance or similarity in this space. A \gls{RKHS} $\mathcal{H}_k$ is a Hilbert space of real-valued functions on $\mathcal{X}$ with the reproducing property that for all $\xx\in\mathcal{X}$ there exists a unique $k_{\xx} \in \mathcal{H}_k$ such that the \emph{evaluation functional} $E_{\xx}(f) = f(\xx)$ can be represented by taking the inner product of this element $k_{\xx}$ and $f$, formally as:
\begin{equation} \label{eq:reproducing}
    E_{\xx}(f) = \langle k_{\xx},f \rangle_{\mathcal{H}_k}.
\end{equation}

Since $k_{\xx'}\in\mathcal{H}_k$ for any $\xx'\in\mathcal{X}$, we can define a kernel function $k(\xx,\xx'):\mathcal{X}\times\mathcal{X}\rightarrow \mathbf{R}$ by letting
\begin{equation}
    k(\xx,\xx')=k_{\xx'}(\xx)=\langle k_{\xx},k_{\xx'} \rangle_{\mathcal{H}_k}.
\end{equation}
Using properties of inner product, it is easy to show that the kernel function $k(\xx,\xx')$ is symmetric and positive definite, and we call it the \emph{reproducing kernel} of the Hilbert space $\mathcal{H}_k$.

\subsection{Functional Gradients} \label{sub:fg}

\emph{Functional derivative} can be thought of as describing the rate of change of the output with respect to the input in a functional. Formally, functional derivative at point $f$ in the direction of $g$ is defined as:
\begin{equation}
  \frac{\partial{L}}{\partial f}(g) = \lim_{\epsilon\rightarrow 0} \frac{L(f+\epsilon g)-L(f)}{\epsilon},
\end{equation}
which is a function of $g$. This is known as \emph{Fréchet derivative} in a Banach space, of which the Hilbert space is a special case.

\emph{Functional gradient}, denoted as $\nabla_{f}L$, is related to functional derivative by the following equation:
\begin{equation}
    \frac{\partial{L}}{\partial f}(g) = \langle \nabla_{f}L, g \rangle_{\mathcal{H}_k}.
\end{equation}

Thanks to the reproducing property, it is straightforward to calculate functional derivative of an evaluation functional in \gls{RKHS}:
\begin{align}
    E_{\xx}(f+\epsilon g) &= \langle f+\epsilon g, k_{\xx} \rangle_{\mathcal{H}_k} \nonumber \\
    &= \langle f, k_{\xx} \rangle_{\mathcal{H}_k} + \epsilon \langle g, k_{\xx} \rangle_{\mathcal{H}_k} \\ 
    \frac{\partial E_{\xx}}{\partial f}(g) &= \langle k_{\xx}, g \rangle_{\mathcal{H}_k}
\end{align}
Therefore, the functional gradient of an evaluation functional is:
\begin{align}
    \nabla_f E_{\xx} &= k_{\xx}.
\end{align}

For a learning task with loss function $\ell$ and a context set $\{(\xx_i,\yy_i)\}_{i\in \CC}$, the overall supervised loss on the context can be written as:
\begin{equation} \label{eq:loss-appendix}
    L(f) = \sum_{i\in \CC} \ell(f(\xx_i),\yy_i).
\end{equation}
In this case, the functional gradient of $L$ can be easily calculated by applying the chain rule:
\begin{align}
    \nabla_f L &= \sum_{i\in \CC} \ell'(f(\xx_i),\yy_i) k_{\xx_i} \\
    &= \sum_{i\in \CC} k(\cdot,\xx_i) \ell'(f(\xx_i),\yy_i).
\end{align}

\subsection{Functional Gradient Descent} \label{sub:fgd-appendix}

To optimise the overall loss on the entire context in \Cref{eq:loss-appendix}, we choose a suitable learning rate $\alpha$, and iteratively update $f$ with:
\begin{align}
    f^{(t+1)}(\xx) &= f^{(t)}(\xx) - \alpha \nabla_f L(f^{(t)})(\xx) \\ 
    &= f^{(t)}(\xx) - \alpha \sum_{i\in \CC} k(\xx,\xx_i) \ell'(f^{(t)}(\xx_i),\yy_i)
\end{align}

In order to evaluate the final model $f^T(\xx)$ at iteration $T$, we only need to compute
\begin{align} \label{eq:f_at_iteration_T}
    f^{(T)}(\xx) = f^{(0)}(\xx) - \sum_{t=0}^{T-1} \alpha \sum_{i\in \CC} k(\xx,\xx_i) \ell'(f^{(t)}(\xx_i),\yy_i),
\end{align}
which does not depend on function values outside the context from previous iterations $t<T$.

\newcommand{\randint}{\fontfamily{pcr}\selectfont randint}
\newcommand{\uniform}{\fontfamily{pcr}\selectfont uniform}
\newcommand{\numpy}{\fontfamily{pcr}\selectfont numpy.random}

\begin{table}[!htb] 
  \centering
  \begin{threeparttable}[]
  \caption{Considered Range of Hyperparameters. The random generators such as {\randint} or {\uniform} use {\numpy} syntax, so the first argument is inclusive while the second argument is exclusive. Whenever a list is given, it means uniformly sampling from the list. $u_{+}$ and $u_{-}$ will be followed by a linear transformation with an output dimension of \emph{dim-reprs}.}
  \begin{tabular}{l|c}
    \toprule
    Components &         Architecture \\ 
    \midrule
    Shared MLP $m$ & \emph{nn-sizes} $\times$ \emph{nn-layers} \\
    MLP for positive labels $u_{+}$ & \emph{nn-sizes} $\times$ \emph{nn-layers} \\
    MLP for negative labels $u_{-}$ & \emph{nn-sizes} $\times$ \emph{nn-layers} \\
    Key/query transformation MLP $a$ & \emph{dim}$(\xx)$ $\times$ \emph{embedding-layers} \\
    Decoder & linear with output dimension \emph{dim}$(\xx)$ \\
    \bottomrule
  \end{tabular}
  \begin{tabular}{l|c}
    \toprule
    Hyperparameters &         Considered Range \\ 
    \midrule
    \emph{num-iters} & \randint(2, 7)  \\
    \emph{nn-layers} & \randint(2, 4)\\
    \emph{embedding-layers} & \randint(1, 3)\\
    \emph{nn-sizes} & $[64,128]$ \\
    \emph{dim-reprs} &  $=$\emph{nn-sizes} \\
    Initial representation $\rr^{0}$ \;\;\;\;\;\;\;\;\;\;\;\;\;\;\; & [zero, constant, parametric] \\ 
    \midrule 
    Outer learning rate & $10^{-5} \times \text{\uniform(-5, -4)}$  \\ 
    Initial inner learning rate & $[0.1, 1.0, 10.0]$ \\ 
    Dropout rate & \uniform(0.0, 0.5) \\ 
    Orthogonality penalty weight & $10^{\text{\uniform(-4, -2)}}$ \\ 
    L2 penalty weight & $10^{\text{\uniform(-10, -8)}}$ \\ 
    Label smoothing & $[0.0, 0.1, 0.2]$ \\ 
    \bottomrule
  \end{tabular}
  \label{tb:hyperparameters-considered}
  \end{threeparttable}
\end{table}

\section{Experimental Details} \label{sec:experimental-details}

We run experiments on Nvidia's GeForce GTX 1080 Ti, and it typically takes about $20$--$40$ minutes to train a few-shot model on a single GPU card until early-stopping is triggered (after seeing $10k$--$100k$ tasks).
For miniImageNet and tieredImageNet, we conduct randomised hyperparameters search \citep{bergstra2012random} for hyperparameters tunning. Typically, $64$ configurations of hyperparameters are sampled for each problem, and the best configuration is chosen by comparing accuracy on the validation set. The considered range of hyperparameters is given in \Cref{tb:hyperparameters-considered}, and the chosen hyperparameters are shown in \Cref{tb:hyperparameters-chosen}. For regression tasks, we simply use hyperparameters listed in \Cref{tb:hyperparameters-regression} for both MetaFun-DFP and MetaFun-KFP.

\begin{table*}[!htb] 
  \centering
  \footnotesize
  \begin{threeparttable}[]
  \caption{Results of randomised hyperparameters search. Hyperparameters shown in this table are not guaranteed to be optimal within the considered range, because we conduct randomised hyperparameters search. However, models configured with these hyperparameters perform reasonably well, and we used them to report final results comparing to other methods. Furthermore, dropout is only applied to the inputs. Orthogonality penalty weight and L2 penalty weight are used in exactly the same way as in \citet{rusu2018meta}. Inner learning rate $\alpha$ is trainable so only an initial inner learning rate is given in the table.}
  \begin{tabular}{l|cc|cc}
  \toprule
     & \multicolumn{2}{c|}{\textbf{miniImageNet}} &  \multicolumn{2}{c}{\textbf{tieredImageNet}}\\ 
    \toprule
    Hyperparameters (for MetaFun-DFP)  &    $1$-shot & $5$-shot & $1$-shot & $5$-shot \\ 
    \midrule
    \emph{num-iters} & $2$ & $5$ & $3$ & $5$ \\
    \emph{nn-layers}  & $3$ & $2$ & $2$ & $3$ \\
    \emph{embedding-layers} & $2$ & $2$ & $1$ & $1$ \\
    \emph{nn-sizes} & $64$ & $128$ & $128$ & $128$ \\
    Initial state & zero & constant & constant & constant \\ 
    \midrule 
    Outer learning rate & $8.56\times10^{-5}$ & $3.71\times10^{-5}$ & $5.55\times10^{-5}$ & $5.78\times10^{-5}$ \\ 
    Initial inner learning rate & $0.1$ & $10.0$ & $1.0$ & $1.0$ \\ 
    Dropout rate & $0.397$ & $0.075$ & $0.123$ & $0.223$ \\ 
    Orthogonality penalty weight & $3.28\times10^{-3}$ & $1.56\times10^{-3}$ &  $1.37\times10^{-3}$ & $2.58\times10^{-3}$ \\ 
    L2 penalty weight & $1.32\times10^{-10}$ & $2.60\times10^{-10}$ & $1.92\times10^{-9}$ & $1.63\times10^{-9}$ \\ 
    Label smoothing & $0.2$ & $0.2$ & $0.1$ & $0.0$ \\ 
    \bottomrule
  \end{tabular}
  
  \begin{tabular}{l|cc|cc}
  \toprule
    &   \multicolumn{2}{c|}{\textbf{miniImageNet}} &  \multicolumn{2}{c}{\textbf{tieredImageNet}}\\ 
    \toprule
    Hyperparameters (for MetaFun-KFP)\;\;\;\;\;  &   $1$-shot & $5$-shot & $1$-shot & $5$-shot \\ 
    \midrule
    \emph{num-iters} & $3$ & $6$ & $4$ & $4$ \\
    \emph{nn-layers} & $3$ & $2$ & $2$ & $3$ \\
    \emph{embedding-layers} & $2$ & $2$ & $1$ & $1$ \\
    \emph{nn-sizes} & $64$ & $64$ & $64$ & $128$ \\
    Initial state & zero & parametric & parametric & zero \\ 
    \midrule 
    Outer learning rate & $4.21\times10^{-5}$ & $8.60\times10^{-5}$ & $8.01\times10^{-5}$ & $4.50\times10^{-5}$ \\ 
    Initial inner learning rate & $0.1$ & $0.1$ & $0.1$ & $0.1$ \\ 
    Dropout rate & $0.424$ & $0.359$ & $0.115$ & $0.148$ \\ 
    Orthogonality penalty weight & $2.69\times10^{-3}$ & $2.73\times10^{-4}$ &  $1.06\times10^{-4}$ & $7.33\times10^{-3}$ \\ 
    L2 penalty weight & $1.19\times10^{-9}$ & $1.68\times10^{-9}$ & $4.90\times10^{-9}$ & $6.22\times10^{-9}$ \\ 
    Label smoothing & $0.2$ & $0.2$ & $0.1$ & $0.1$ \\ 
    \bottomrule
  \end{tabular}
  \label{tb:hyperparameters-chosen}
  \end{threeparttable}
\end{table*}

\begin{table*}[!htb] 
  \centering
  \begin{threeparttable}[]
  \caption{Hyperparameters for regression tasks. Local update function and the predictive model will be followed by linear transformations with output dimension of \emph{dim-reprs} and \emph{dim}(\yy) accordingly.}
  \begin{tabular}{l|c}
    \toprule
    Components &         Architecture \\ 
    \midrule
    Local update function & \emph{nn-sizes} $\times$ \emph{nn-layers} \\
    Key/query transformation MLP $a$ & \emph{nn-sizes} $\times$ \emph{embedding-layers} \\
    Decoder & \emph{nn-sizes} $\times$ \emph{nn-layers} \\
    Predictive model & \emph{nn-sizes} $\times$ (\emph{nn-layers}-1) \\
    \bottomrule
  \end{tabular}
  \begin{tabular}{l|c}
    \toprule
    Hyperparameters &         Considered Range \\ 
    \midrule
    \emph{num-iters} & $5$ \\
    \emph{nn-layers} & $3$\\
    \emph{embedding-layers} & $3$ \\
    \emph{nn-sizes} & $128$ \\
    \emph{dim-reprs} &  $=$\emph{nn-sizes} \\
    Initial representation $\rr^{0}$ \;\;\;\;\;\;\;\;\;\;\;\;\;\;\; & zero \\ 
    \midrule 
    Outer learning rate & $10^{-4}$  \\ 
    Initial inner learning rate & $0.1$ \\ 
    Dropout rate & $0.0$ \\ 
    Orthogonality penalty weight & $0.0$ \\ 
    L2 penalty weight & $0.0$ \\ 
    \bottomrule
  \end{tabular}
  \label{tb:hyperparameters-regression}
  \end{threeparttable}
\end{table*}

\end{document}